\documentclass[letterpaper, 10 pt, conference]{ieeeconf}  

\IEEEoverridecommandlockouts                              

\usepackage{amsmath} 
\usepackage{amssymb}  
\usepackage{bm}
\usepackage{url}
\usepackage{booktabs}
\usepackage{caption}
\usepackage{tabularx}
\usepackage{multirow}
\usepackage{graphicx}
\usepackage{xcolor}
\usepackage{subcaption}

\title{\LARGE \bf
RigPI: Dynamic Parameter Identification of Rigid Body via VLM-Seeded Differentiable Simulation
}

\author{
Xincheng He$^{1}$,
Rongrong Zhang$^{2}$,
Wei Jiang$^{1}$,
Wenqiang Xu$^{3}$*%
\thanks{$^{1}${\tt\small \{xinchenghe, ailon\_jw\}@sjtu.edu.cn}.}%
\thanks{$^{2}${\tt\small zhangrongrong@way-robotics.com}}%
\thanks{$^{3*}${\tt\small vinjohn@nus.edu.sg}. Wenqiang Xu is the corresponding author.}
}

\begin{document}

\maketitle
\thispagestyle{empty}
\pagestyle{empty}

\begin{abstract}
Accurate physical parameter identification of manipulated objects is fundamental to advanced robotic manipulation and the construction of faithful digital twins. However, acquiring physically consistent inertial and frictional properties from real-world interactions remains challenging due to sensing noise, modeling errors, and limited prior knowledge. This paper presents \textbf{RigPI}, a systematic framework for identifying dynamic parameters of both unconstrained rigid bodies and multi-link rigid bodies during robot–object interaction. RigPI integrates vision-based semantic priors, force–torque measurements, and motion observations within a differentiable simulation pipeline. A vision-language model (VLM) provides informed initialization and a constrained search space, while gradient information from a differentiable physics simulator enables efficient and stable parameter refinement. The proposed two-stage optimization strategy alleviates sensitivity to noise and avoids physically implausible solutions. Extensive real-world experiments on objects with revolute and prismatic joints demonstrate that RigPI achieves accurate and stable parameter estimates, and successfully reproduces manipulation trajectories on a real robot with parameter-aware predictive validity. These results highlight the effectiveness and robustness of RigPI for real-world robotic system identification tasks. For code, model, and more supporting materials, please visit our website at \url{https://sites.google.com/view/rigpi}.

\end{abstract}

\section{INTRODUCTION}

Accurate dynamic parameter identification of external objects, including both inertial and frictional properties, is essential for advanced robotic manipulation, such as predictive control, and creating digital twins that mirror real-world physics. However, accurate and automatic acquisition of such parameters during manipulation remains a significant challenge. It requires leveraging diverse sources of information from observed interaction data, such as trajectories, force/torque signals, and prior knowledge, while optimizing the parameters in a carefully constrained search space to avoid physically implausible solutions. To this end, we introduce RigPI, which integrates high-level priors with rich mechanical feedback and orchestrates them within a gradient-based optimization pipeline.

Traditional approaches to dynamic parameter identification are typically based on inverse-dynamics formulations, where inertial parameters are estimated as linear functions of joint torques using least-squares or robust variants \cite{4048449,4650672,Xu_Fan_Fang_Zhu_Zhao_2022}. These methods assume precise knowledge of system dynamics and are highly sensitive to measurement noise, which directly corrupts both the regressor and the observation vector in the linear system. They also require carefully designed excitation trajectories to render the regression problem well-posed. As a result, small modeling inaccuracies or sensing disturbances may lead to physically inconsistent or unstable parameter estimates. To mitigate these challenges, recent developments have shifted attention toward differentiable simulation-based identification schemes that evaluate candidate parameters through forward simulation of the system’s time evolution. Recent differentiable physics simulators enable gradient-based system identification in complex, contact-rich environments \cite{newton,hu2019difftaichi,howelllecleach2022}. They provide powerful solvers for parameter optimization and offer rich support for geometry, collision, and other environmental features \cite{lidec2021,gradsim}.

In this work, we propose \textbf{RigPI}, a real-to-sim-to-real framework that tackles \textbf{rigid} object identification through a systematic process: First, it takes as input RGB-D images of the object and robot–object interaction data (applied forces, torques, and resulting poses), and then aligns the object state from the real world to a differentiable simulator. Second, it adopts a two-stage optimization mechanism to estimate the object’s dynamic parameter vector $\bm{\theta}$ through the differentiable simulator. A vision-language model (VLM) is adopted to produce an informed initialization and a feasible search space from images, and then differentiable simulation with NVIDIA Newton \cite{newton} provides gradients that drive gradient-based optimization to refine the parameters efficiently.

\begin{figure}[t]
    \centering
    \includegraphics[width=\columnwidth]{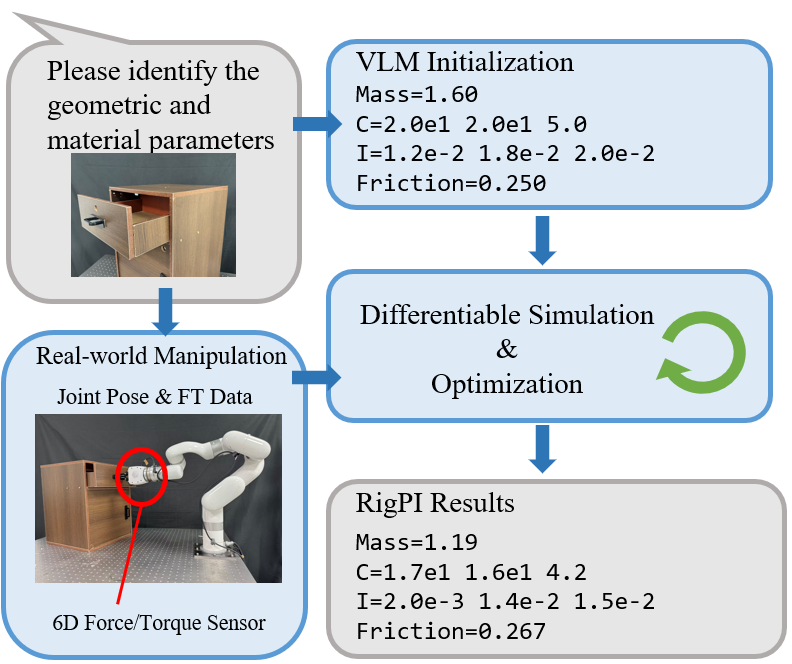}
    \caption{RigPI estimates an object’s mass, inertia, and friction properties during manipulation.}
    \label{fig:overview}
\end{figure}

To evaluate the proposed pipeline, we conduct a comprehensive test on a real-world robotic platform using a diverse set of external objects, including both unconstrained rigid bodies and articulated bodies with revolute and prismatic joints. We assess the accuracy and stability of the identified parameters and further validate their correctness by reproducing manipulation trajectories on real hardware using the parameter-aware predictive validity. Our results demonstrate that RigPI can achieve high-fidelity parameter identification while ensuring stability in real-world robotic applications. In addition, we perform ablation studies to verify the accuracy and necessity of the visual priors in guiding the identification process.

We summarize our contributions as follows:
\begin{itemize}
    \item We present RigPI, a pipeline to identify the dynamic parameters of both unconstrained and multi-link rigid bodies, including inertial and frictional properties, via robotic manipulation.
    \item We conduct real-world robot tasks and evaluations to demonstrate its practical applicability and superior performance on both unconstrained and multi-link rigid objects.
\end{itemize}

\section{RELATED WORKS}
\label{sec:related works}

\subsection{Differentiable Simulation for Parameter Identification}

Traditional parameter identification is largely based on inverse-dynamics formulations, where joint torques are modeled as linear functions of inertial parameters. Classical work by Atkeson et al. \cite{4048449} established a least-squares framework for identifying mass, center of mass, and inertia. Subsequent studies improved robustness and adaptability through recursive total least-squares schemes \cite{4650672} and noise-aware weighted least-squares variants \cite{Xu_Fan_Fang_Zhu_Zhao_2022}. Despite their effectiveness in structured settings, these methods require carefully designed excitation trajectories, assume simplified contact and friction models, and are highly sensitive to measurement noise and modeling errors. Moreover, their linear assumptions often fail in contact-rich or visually inferred manipulation scenarios, resulting in ill-posed estimates or physically inconsistent parameters.

With the rise of differentiable simulators, modern methods now leverage analytic gradients of physics engines to perform end-to-end optimization, offering a unified framework that integrates sensor feedback and gradient-based optimization for system identification. Early explorations of differentiable dynamics \cite{Toussaint2018DifferentiablePA} have evolved into general-purpose systems such as \textit{DiffTaichi} \cite{hu2019difftaichi}, as well as physically accurate and highly optimized engines like \textit{Dojo} \cite{howelllecleach2022} and \textit{Newton} \cite{newton}. These GPU-accelerated simulators provide efficient forward and backward passes for gradient-based optimization. However, they are primarily computation backbones rather than complete identification solutions: performance remains sensitive to initialization and parameter bounds, and they lack mechanisms for constructing well-posed problems from unstructured real-world data.

Recent studies have advanced differentiable physics simulation by enabling optimization over high-dimensional or neural-augmented objects \cite{lidec2021,le_cleach_differentiable_2023,degrave2018differentiablephysicsenginedeep}, integrating deep learning for robotics applications, and providing systematic frameworks for parameter optimization and model fitting. In robotic system identification, these methods have predominantly focused on inertial parameters of single, monolithic rigid bodies \cite{chen2025learningobjectpropertiesusing}, or employed data-driven models such as Graph Networks \cite{sanchezgonzalez2018graphnetworkslearnablephysics} and physics-informed Deep Lagrangian Networks \cite{lutter2019deeplagrangiannetworksusing}. Other works have extended differentiable parameter identification to fluids and deformable objects, e.g., DiffStir \cite{diffstir} and DiffCP \cite{diffcp}.
While these approaches offer complementary strategies, they do not fully address the systematic identification of unconstrained and multi-link bodies with physics-grounded, interpretable dynamic parameters.

Our framework, \textbf{RigPI}, bridges this gap by formulating the identification problem as a structured two-stage process. In contrast to purely learning-based methods that operate as black boxes, RigPI preserves the physical interpretability of the estimated parameters ($\bm{\theta}$), ensuring both accuracy and transparency.

\subsection{Object-Centric Real-World System Identification}

Recent progress in foundation models has revived the idea of inferring physical properties from visual data. Early CNN-based approaches regressed quantities such as mass from simulated interactions \cite{wu2015nips,pmlr-v78-standley17a}, but suffered from limited generalization. The emergence of Vision-Language Models (VLMs) enables zero-shot, common-sense physical reasoning, with benchmarks such as \textit{PhysBench} \cite{chow2025physbench} quantifying these capabilities.

However, current VLM-based systems are not designed for precise real-world system identification. They typically operate in open-loop settings or serve as components within Vision-Language-Action (VLA) pipelines \cite{brohan2023rt2visionlanguageactionmodelstransfer}, which emphasize task planning rather than quantitative parameter estimation. Nevertheless, recent studies demonstrate the potential of physically grounded reasoning, including VLM-based liquid reasoning \cite{lai2024visionlanguagemodelbasedphysicalreasoning}, robotic reachability analysis \cite{zhou2025physvlmenablingvisuallanguage}, and navigation with physical constraints \cite{elnoor2024robotnavigationusingphysically}. These efforts suggest that visual priors can be effectively coupled with physical inference beyond semantic reasoning alone.

RigPI addresses this gap by treating the VLM's output as an informed prior rather than a final answer, formulating an optimization problem that tightly couples object-centric priors with high-fidelity gradient information from a differentiable simulator. This synergy establishes a robust identification framework capable of overcoming the limitations of approaches that rely solely on either semantic priors or differentiable simulation.

\section{PROBLEM STATEMENT}
\label{sec:problem statement} 

The central problem addressed in this research is dynamic parameter identification for \textbf{external objects}, which aims to estimate mass, inertial, and frictional properties by observing their response to a series of controlled manipulations. 


\subsection{Dynamic Parameters}
For a rigid body, the concerned parameters,  which cover major parameters affecting the Newtonian motion are given as follows:
\begin{itemize} 
    \item \textbf{Mass ($m$):} A scalar quantity representing the body's translational inertia. 
    \item \textbf{Center of Mass ($\bm{com}$):} A vector in $\mathbb{R}^3$ specifying the mean position of the mass distribution. 
    \item \textbf{Inertia Tensor ($\bm{I}$):} A $3 \times 3$ symmetric positive-definite matrix characterizing the body's rotational inertia relative to its center of mass.
    \item \textbf{Coefficient of Surface Friction ($\mu$):} A scalar modeling frictional forces arising from contact with external surfaces. 
\end{itemize}

To note, such a definition can be naturally extended to \textbf{multi-link bodies} which have only a single rigid part moved under manipulation along a 1-DoF joint, such as a cabinet or a drawer. 
For the friction property in unconstrained and multi-link rigid bodies, our model adopts Coulomb friction.


To put the parameters together and make them suitable for optimization, we vectorize all identifiable parameters into $\bm{\theta}$:
\begin{equation}
    \bm{\theta} = \left( m, \bm{com}, \bm{I}, \mu \right).
\end{equation}

For an unconstrained rigid body, this denotes the vectorized unique elements of mass, center of mass, inertia tensor, and friction coefficient. For a multi-link rigid body, this vector is simply extended to encompass the full set of parameters $\bm{\theta}_{i,j}=(m_i, \bm{com}_i, \bm{I}_i, \mu_{\text{joint},j})$ from all its constituent links and joints, with $i$ and $j$ respectively indexing the links and joints in the articulated structure.

\subsection{Parameter Identification}
Ultimately, our goal is to find an optimal parameter vector $\bm{\theta}^*$ that best explains the observed physical behavior. However, searching the entire parameter space is inefficient and prone to physically implausible solutions. Therefore, we formulate this as a constrained optimization problem. This requires both an initial parameter guess, $\bm{\theta}^{(0)}$, and a corresponding constraint for optimization, $\mathcal{C}$, to define the feasible search space. The initial guess is a vector of estimated parameters:
\begin{equation}
    \bm{\theta}^{(0)} = \left(\hat{m}_0, \widehat{\bm{com}}_0, \widehat{\bm{I}}_0, \hat{\mu}_0 \right).
\end{equation}
The optimization is then confined within the range:
\begin{equation}
    \mathcal{C} = [\bm{\theta}_{\min}, \bm{\theta}_{\max}].
\end{equation}
Both the initial guess and the plausible range are derived from high-level base properties (e.g., density, dimensions) inferred from the object's visual appearance, as summarized in Table~\ref{tab:vlm_params}.

\begin{table}[h!]
\centering
\caption{Parameter constraints derived from visual priors.}
\label{tab:vlm_params}
\begin{tabular}{@{}ll@{}}
\toprule
\textbf{Parameter} & \textbf{Derived Constraint} \\
\midrule
Mass ($m$) & Plausible range $[m_{\min}, m_{\max}]$ \\
Center of Mass ($\bm{com}$) & Plausible range $[\bm{com}_{\min}, \bm{com}_{\max}]$ \\
Inertia Tensor ($\bm{I}$) & Element-wise bounds $[\bm{I}_{\min}, \bm{I}_{\max}]$ \\
Friction Coeff. ($\mu$) & Plausible range $[\mu_{\min}, \mu_{\max}]$ \\
\bottomrule
\end{tabular}
\end{table}

The final optimization objective is thus to find the parameter vector $\bm{\theta}^*$ that minimizes a loss function $\mathcal{L}(\bm{\theta})$ while remaining within this physically-grounded search space, starting from the initial guess $\bm{\theta}^{(0)}$. The specific methodology for deriving the initial values and bounds, and for formulating the loss function, will be detailed in Section~\ref{sec:method}.

\begin{figure*}[htbp]
    \centering
    \includegraphics[width=0.85\textwidth]{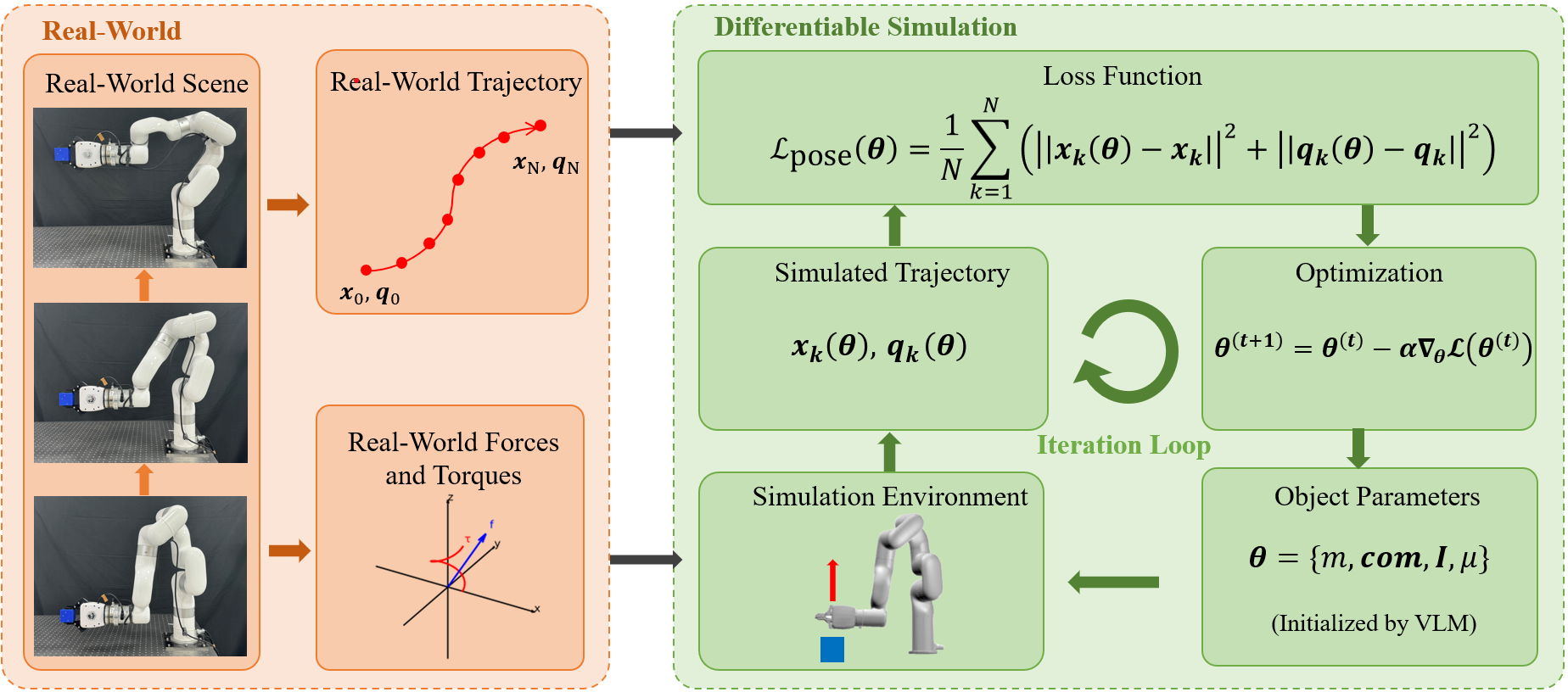}
    \caption{RigPI takes recorded robot interaction data as input, including forces, torques, and poses, and is initialized with a parameter vector, $\bm{\theta}^{(0)}$, provided by a VLM. This initiates an iterative optimization loop within a differentiable simulator. In each iteration, the simulator performs a forward pass, using the current parameter guess and the measured real-world forces to generate a simulated trajectory. A loss function then quantifies the discrepancy between this simulated trajectory and the ground-truth trajectory. Leveraging automatic differentiation, the framework computes the gradient of this loss with respect to the parameters. Finally, a gradient-based optimizer uses this gradient to update the parameters, iteratively refining them to minimize the simulation-to-reality gap and yield the final, accurate model.}
    \label{fig:pipeline}
\end{figure*}

\section{RigPI METHOD}
\label{sec:method}

Our proposed RigPI addresses the challenge of identifying the physical parameters of rigid parts through a real-to-sim-to-real, differentiable optimization framework. The approach first acquires data from the real world in Sec. \ref{subsec:data_acquisition}, and then performs parameter optimization in Sec. \ref{subsec:sim_and_opt} with vision-guided parameter initialization and gradient-based optimization. The overall pipeline of RigPI is depicted in Fig.~\ref{fig:pipeline}.

\subsection{Object State Observation}
\label{subsec:data_acquisition}

In the initial stage, we acquire experimental data by having a robotic manipulator interact with the object of interest, which can be either an \textbf{unconstrained rigid body} or a rigid part of a \textbf{multi-link body}. The robot records two primary types of information for each interaction:

\begin{itemize}
    \item \textbf{Wrench ($\bm{f}, \bm{\tau}$):} This represents the forces and torques applied to the object under test during manipulation. The interaction is captured by a force vector $\bm{f} \in \mathbb{R}^3$ and a torque vector $\bm{\tau} \in \mathbb{R}^3$, which is applied to the center of mass of the object being identified.

    \item \textbf{Object Pose ($\bm{x}, \bm{q}$):} This describes the position and orientation of the object being tracked. The position of its center of mass is given by the vector $\bm{x} \in \mathbb{R}^3$. Its orientation is represented by a unit quaternion $\bm{q} \in SO(3)$. For a multi-link rigid body, although the whole state includes all joint angles, we focus only on the interacting rigid parts here.

\end{itemize}

The collected dataset, denoted as $\mathcal{D} = \{(\bm{f}_k, \bm{\tau}_k, \bm{x}_k, \bm{q}_k)\}_{k=1}^N$, where $k$ means the $k$-th iteration step and $N$ means the number of iteration steps, providing a comprehensive record of the object's dynamic behavior under controlled robotic manipulation.

\subsection{Parameter Initialization and Optimization}
\label{subsec:sim_and_opt}

After collecting interaction data, our next step is to identify the object's dynamic parameters by minimizing a loss function that measures the discrepancy between simulated and observed behavior. Since this optimization is highly sensitive to the initial guess, obtaining a good initialization is crucial. To this end, we leverage Vision-Language Models (VLMs), such as Intern-S1 \cite{bai2025interns1scientificmultimodalfoundation} and Gemini-3 Pro \cite{comanici2025gemini25pushingfrontier}, to extract an informed initial parameter vector $\bm{\theta}^{(0)}$ and define a plausible search space $\mathcal{C}$. Specifically, we input clear and unoccluded RGB-D images of the target object into the VLM, which can infer its geometric properties (pose, scale, volume, internal structure) and material characteristics, providing initial estimates and bounds for density ($\rho$) and friction coefficient ($\mu$), as defined in Section~\ref{sec:problem statement}. Combined with the assumption of uniform mass distribution, this yields a full set of initial parameters and their corresponding bounds.

With the initialized parameter, the identification process can be done via differentiable simulation using the NVIDIA Newton physics engine \cite{newton}. The collected wrench data $(\bm{f}_k, \bm{\tau}_k)$ drive the simulation, which propagates the object's state over time given the current parameter guess $\bm{\theta}$. For unconstrained rigid bodies, a semi-implicit integrator is used, whereas multi-link rigid bodies are simulated using a Featherstone integrator.

\begin{figure*}[t!]
    \centering
    \definecolor{predictedcolor}{RGB}{0, 0, 255}

    \begin{subfigure}[t]{0.19\textwidth}
        \centering
        \includegraphics[width=\linewidth,height=3cm,keepaspectratio]{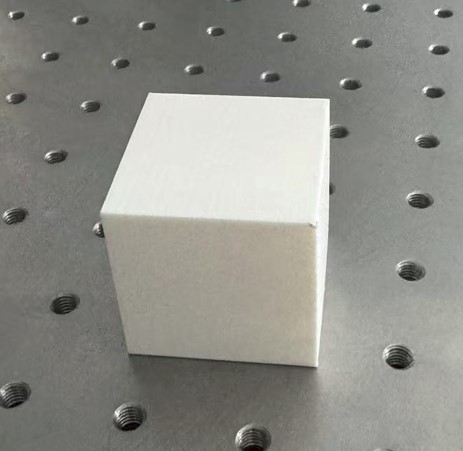}
        \caption{Cube}
    \end{subfigure}%
    \hfill
    \begin{subfigure}[t]{0.19\textwidth}
        \centering
        \includegraphics[width=\linewidth,height=3cm,keepaspectratio]{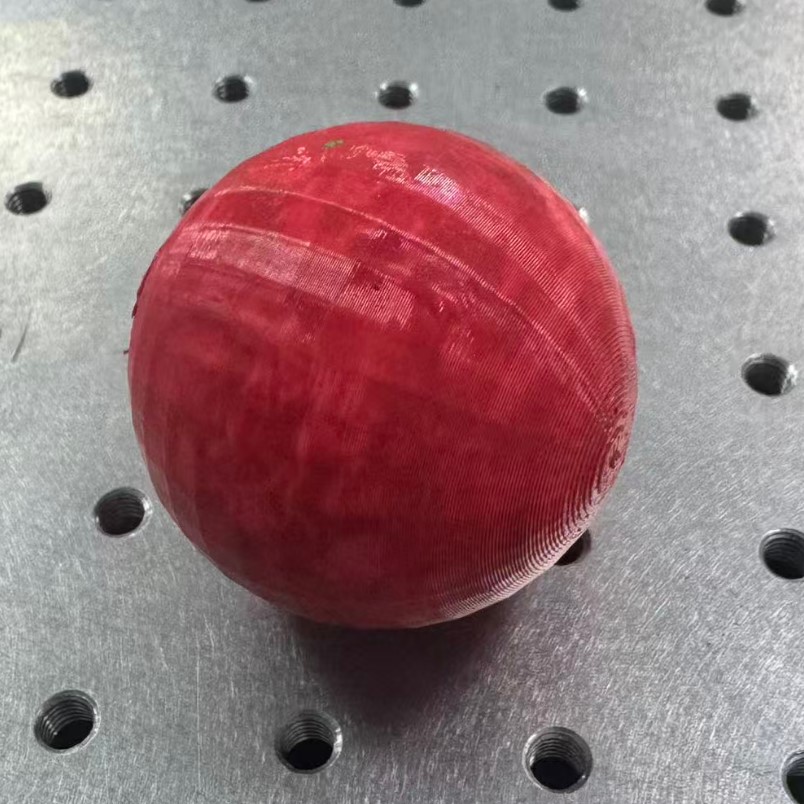}
        \caption{Sphere}
    \end{subfigure}%
    \hfill
    \begin{subfigure}[t]{0.19\textwidth}
        \centering
        \includegraphics[width=\linewidth,height=3cm,keepaspectratio]{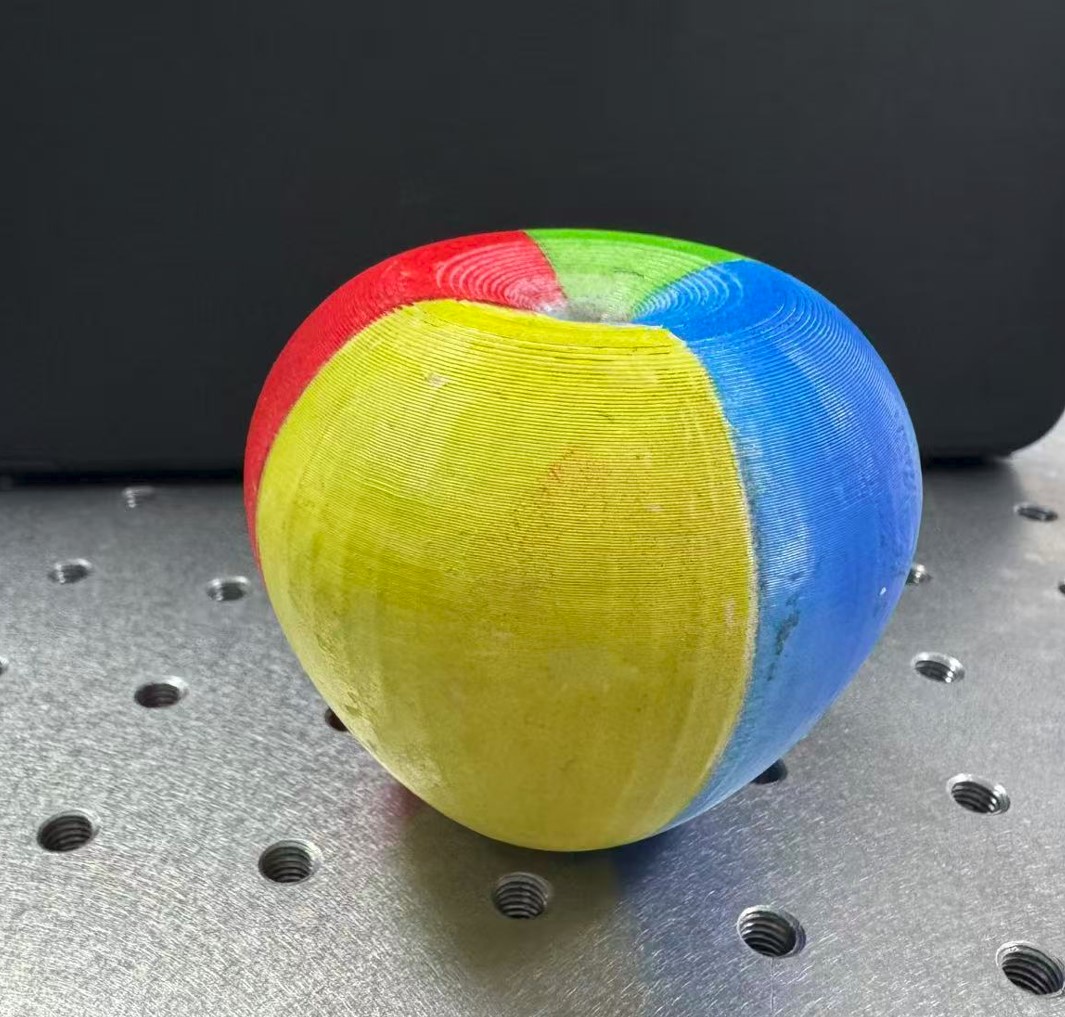}
        \caption{Apple}
    \end{subfigure}%
    \hfill
    \begin{subfigure}[t]{0.19\textwidth}
        \centering
        \includegraphics[width=\linewidth,height=3cm]{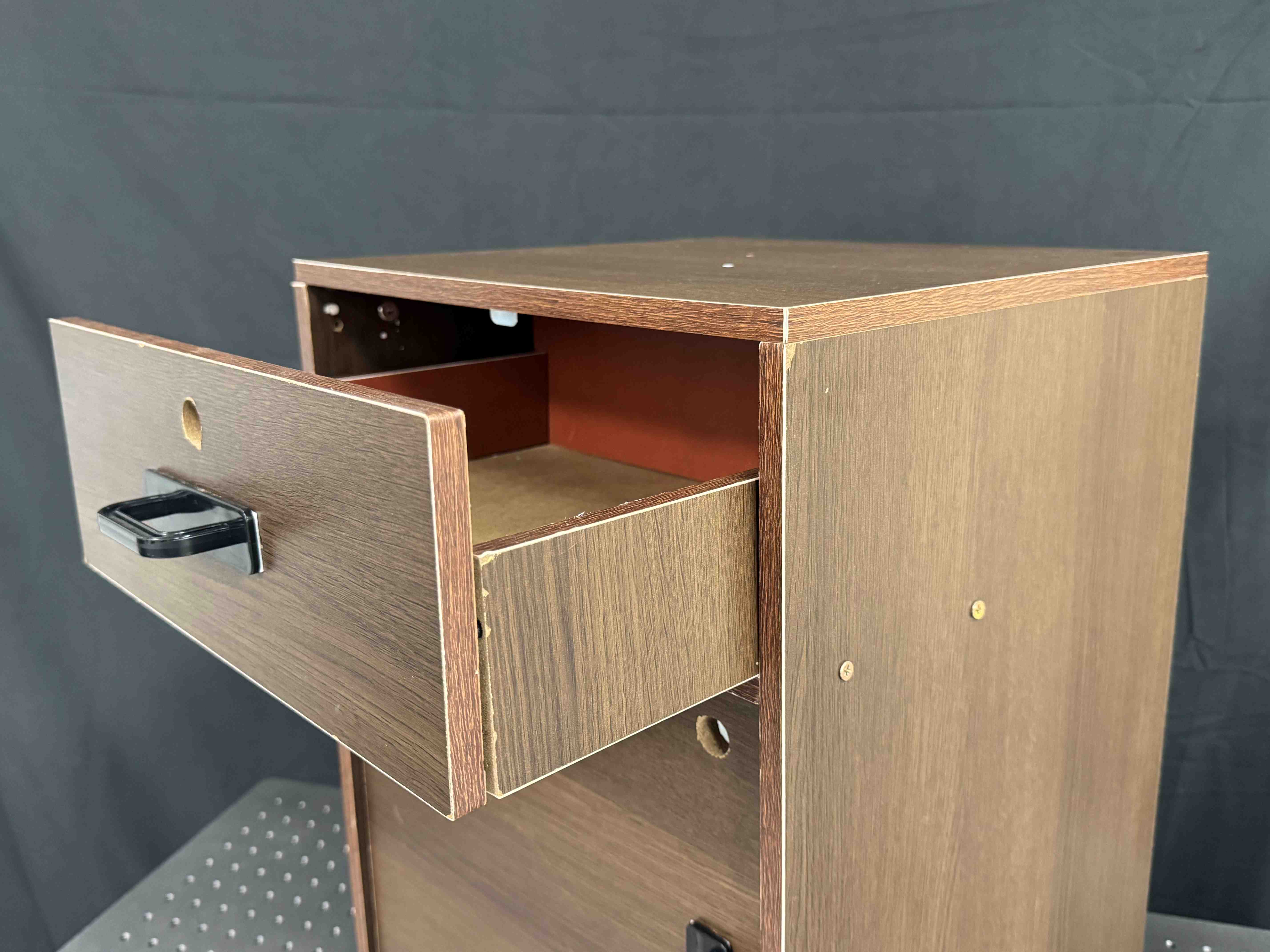}
        \caption{Drawer}
    \end{subfigure}%
    \hfill
    \begin{subfigure}[t]{0.19\textwidth}
        \centering
        \includegraphics[width=\linewidth,height=3cm]{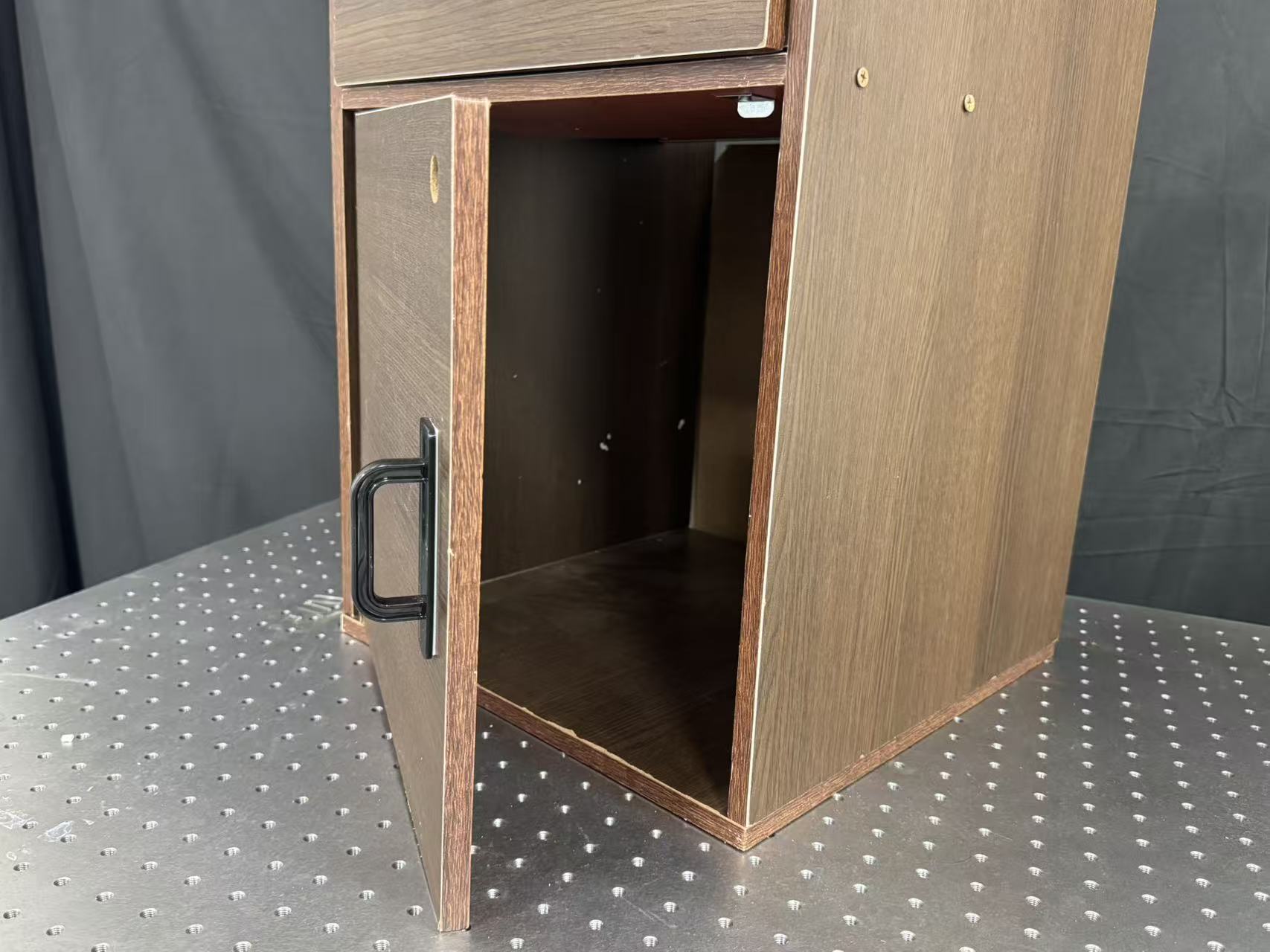} 
        \caption{Door}
    \end{subfigure}

    \vspace{0.5em}

    \begin{subfigure}[t]{0.19\textwidth}
        \centering
        \small
        Mass=\textcolor{predictedcolor}{2.03e-2} \\
        $\boldsymbol{C}$=\textcolor{predictedcolor}{[1.4e-2 2.5e-2 1.4e-2]} \\
        $\boldsymbol{I}$=\textcolor{predictedcolor}{[5.8e-6 5.7e-6 5.7e-6]} \\
        Friction=\textcolor{predictedcolor}{0.386}
    \end{subfigure}%
    \hfill
    \begin{subfigure}[t]{0.19\textwidth}
        \centering
        \small
        Mass=\textcolor{predictedcolor}{2.07e-2} \\
        $\boldsymbol{C}$=\textcolor{predictedcolor}{[2.8e-2 3.1e-2 1.6e-2]} \\
        $\boldsymbol{I}$=\textcolor{predictedcolor}{[5.3e-6 5.2e-6 5.4e-6]} \\
        Friction=\textcolor{predictedcolor}{0.356}
    \end{subfigure}%
    \hfill
    \begin{subfigure}[t]{0.19\textwidth}
        \centering
        \small
        Mass=\textcolor{predictedcolor}{4.23e-2} \\
        $\boldsymbol{C}$=\textcolor{predictedcolor}{[2.1e-2 1.8e-2 1.4e-1] } \\
        $\boldsymbol{I}$=\textcolor{predictedcolor}{[1.4e-5 1.3e-5 1.3e-5] } \\
        Friction=\textcolor{predictedcolor}{0.377}
    \end{subfigure}
    \hfill
    \begin{subfigure}[t]{0.19\textwidth}
        \centering
        \small
        Mass=\textcolor{predictedcolor}{1.19} \\
        $\boldsymbol{C}$=\textcolor{predictedcolor}{[1.7e1 1.6e1 4.2]} \\
        $\boldsymbol{I}$=\textcolor{predictedcolor}{1.4e-3 1.4e-2 1.2e-2]} \\
        Friction=\textcolor{predictedcolor}{0.287}
    \end{subfigure}%
    \hfill
    \begin{subfigure}[t]{0.19\textwidth}
        \centering
        \small
        Mass=\textcolor{predictedcolor}{8.51e-1} \\
        $\boldsymbol{C}$=\textcolor{predictedcolor}{[1.6e1 3.5e-1 6.2e-1]} \\
        $\boldsymbol{I}$=\textcolor{predictedcolor}{[7.7e-3 8.4e-3 1.9e-2]} \\
        Friction=\textcolor{predictedcolor}{0.429}
    \end{subfigure}

    \caption{Examples of physical parameter identification using our RigPI system. Each subfigure shows the unconstrained or multi-link rigid body being identified (left) and the predicted values. Units are: Mass [kg], $\boldsymbol{C}$ [m], $\boldsymbol{I}$ [kg$\cdot$m$^2$], Friction (coefficient of friction, unitless). $\boldsymbol{C}$ denotes center of mass, and $\boldsymbol{I}$ denotes rotational inertia.}
    \label{fig:RigPI_results}
    \vspace{-0.5cm}
\end{figure*}

To guide the optimization, we define an augmented \textbf{loss function} that combines a pose discrepancy term with a penalty for violating physical constraints. The primary loss term, $\mathcal{L}_{\text{pose}}(\bm{\theta})$, quantifies the difference between the simulated and empirically observed poses:
\begin{equation}
\label{equa:loss function}
    \mathcal{L}_{\text{pose}}(\bm{\theta}) = \frac{1}{N} \sum_{k=1}^N \left( \|\bm{x}_{k}(\bm{\theta}) - \bm{x}_k\|^2 + \|\bm{q}_{k}(\bm{\theta}) - \bm{q}_k\|^2 \right).
\end{equation}
To this, we add a penalty term, $P(\bm{\theta}, \mathcal{C})$, which enforces the parameter bounds 
$\mathcal{C} = [\bm{\theta}_{\min}, \bm{\theta}_{\max}]$. Specifically, $P(\bm{\theta}, \mathcal{C})$ is defined as
\begin{equation}
\begin{split}
    P(\bm{\theta}, \mathcal{C}) = \sum_{i=1}^{d} \Big( 
    &\max\{0, \theta_i - \theta_{\max,i}\}^2 \\
    + &\max\{0, \theta_{\min,i} - \theta_i\}^2
    \Big),
\end{split}
\end{equation}
where $d$ is the dimension of $\bm{\theta}$. This formulation ensures that no penalty is applied if $\bm{\theta}$ lies within the valid region $\mathcal{C}$, while quadratic penalties are imposed for violations of the lower or upper bounds. The final objective is to minimize the total loss, $\mathcal{L}_{\text{total}}(\bm{\theta})$:
\begin{equation}
    \mathcal{L}_{\text{total}}(\bm{\theta}) = \mathcal{L}_{\text{pose}}(\bm{\theta}) + \lambda P(\bm{\theta}, \mathcal{C}),
\end{equation}
where $\lambda$ is a weighting hyperparameter that controls the strength of the penalty. The differentiable nature of the NVIDIA Newton engine allows for the efficient computation of the gradient $\nabla_{\bm{\theta}} \mathcal{L}_{\text{total}}(\bm{\theta})$.

The final stage employs an iterative gradient descent algorithm to find the optimal parameters $\bm{\theta}^*$. The optimization objective is defined as:
\begin{equation}
    \bm{\theta}^* = \arg\min_{\bm{\theta}} \left( \mathcal{L}_{\text{pose}}(\bm{\theta}) + \lambda P(\bm{\theta}, \mathcal{C}) \right),
\end{equation}
where $\lambda$ is a weighting hyperparameter controlling the penalty strength. In all our experiments, we empirically set $\lambda$ to $0.1$. Starting from the initial guess $\bm{\theta}^{(0)}$ and using an adaptive learning rate, the iterative update rule is given by:
\begin{equation}
    \bm{\theta}^{(t+1)} = \bm{\theta}^{(t)} - \alpha^{(t)} \nabla_{\bm{\theta}} \mathcal{L}_{\text{total}}(\bm{\theta}^{(t)}).
\end{equation}
where $\alpha^{(t)}$ denotes the learning rate (step size) at iteration $t$. The penalty term creates steep gradients that steer the search away from invalid parameter regions defined by the visual priors. This formulation enforces the constraints implicitly, eliminating the need for specialized constrained-optimization solvers.

We scale $\alpha$ according to the magnitude of the initial data. Specifically, at each iteration, the step size is set such that the update remains roughly two orders of magnitude smaller than the original parameter scale, which is defined by the initial parameter and the range given by VLM. This ensures that each optimization step is neither too aggressive to cause instability nor too small to slow convergence, effectively adapting the learning rate to the scale of the parameters being optimized. Further discussion about the comparison between dynamic and fixed is shown in Sec~\ref{subsec:ablation}.

\section{EXPERIMENTAL SETUP}

\subsection{Simulation Setup}
\label{subsec:simulation_setup}

Our differentiable physics simulations were performed using the NVIDIA Newton framework \cite{newton}. All computations were accelerated on an NVIDIA GeForce RTX 4060 GPU, enabling efficient gradient calculations that are crucial to our optimization pipeline. The runtime per optimization iteration is $0.1$ seconds.

\vspace{-0.2cm}
\subsection{Real-World Setup}
\label{subsec:real_world_setup}
All experiments were conducted on a sturdy optical table, providing a stable foundation for all components.

\subsubsection{Robotic Manipulator} 
We used a uFactory xArm6 robotic arm, securely fixed to the table and equipped with a custom gripper. Gripper position and orientation ($\bm{x}, \bm{q}$) were obtained from the xArm6’s internal readings.

\subsubsection{Force/Torque Sensor} 
Force and torque were measured using an ATI Axia80-M20 sensor at the gripper base, with a resolution of ~1/80 N (force) and ~1/2000 N·m (torque), enabling precise contact characterization.

\subsubsection{Object Setup and Data Transformation} 
We used 3D-printed objects for parameter identification: unconstrained rigid bodies (cube, sphere, apple) and two-joint rigid bodies (cabinet, drawer, oven). Objects were first fixed to the table for stability, and all collected data—force, torque, position, orientation—were transformed to the world frame.

\subsubsection{Vision Sensor} 
We use an Azure Kinect camera to capture RGB-D images of the objects, serving as input for VLM estimation and point cloud reconstruction.

\begin{table*}[t!]
\centering
\caption{Accuracy evaluation for unconstrained and multi-link rigid bodies with known ground truth. The table reports ground-truth (GT) parameters, VLM-initialized estimates (with relative error), and the final identified results from our RigPI (with relative error).}
\label{tab:identification_accuracy}
\begin{tabular}{@{}llccccc@{}}
\toprule
\textbf{Object} & \textbf{} & \textbf{Mass (kg)}  & \textbf{$I_{xx}$ (kg$\cdot$m$^2$)} & \textbf{$I_{yy}$} (kg$\cdot$m$^2$) & \textbf{$I_{zz}$ (kg$\cdot$m$^2$)} \\
\midrule
\multirow{4}{*}{\textit{Cube}} 
& Ground Truth & 2.00e-2  & 5.33e-6 & 5.33e-6 & 5.33e-6 \\
& Init(Intern-S1) & 8.00e-2 / 300.00\% & 2.13e-5 / 299.81\% & 2.13e-5 / 299.81\% & 2.13e-5 / 299.81\% \\
& Init(Gemini-3 Pro) & 5.40e-2 / 170.00\% & 1.10e-5 / 106.36\% & 1.10e-5 / 106.36\% & 1.10e-5 / 106.36\% \\
& Least-Square Baseline  &1.72e-2 / 14.00\% & 4.05e-6 / 24.02\% & 4.31e-6 / 19.14\% & 4.09e-6 / 23.26\%   \\
& Gradsim Baseline &1.82e-2 / 9.00\% & 4.29e-6 / 19.51\% & 4.67e-6 / 12.38\% & 4.05e-6 / 24.02\%   \\
& RigPI(Ours) & \textbf{2.03e-2 / 1.50\%} & \textbf{5.79e-6 / 8.63\%} & \textbf{5.67e-6 / 6.38\%} & \textbf{5.71e-6 / 7.13\%} \\
\midrule
\multirow{4}{*}{\textit{Sphere}} 
& Ground Truth & 2.00e-2  & 5.00e-6 & 5.00e-6 & 5.00e-6 \\
& Init(Intern-S1) & 7.85e-2 / 292.50\% & 1.96e-5 / 292.00\% & 1.96e-5 / 292.00\% & 1.96e-5 / 292.00\% \\
& Init(Gemini-3 Pro) & 6.87e-2 / 243.50\% & 1.73e-5 / 246.00\% & 1.73e-5 / 246.00\% & 1.73e-5 / 246.00\% \\
& Least-Square Baseline &1.78e-2 / 11.00\% & 3.69e-6 / 26.20\% & 4.21e-6 / 15.80\% & 3.72e-6 / 17.80\%  \\
& Gradsim Baseline &19e-2 / 9.50\% & 5.88e-6 / 17.60\% & 4.31e-6 / 13.80\% & 5.89e-6 / 17.80\%  \\
& RigPI(Ours) & \textbf{2.07e-2 / 3.50\%} & \textbf{5.37e-6 / 7.40\%} & \textbf{5.21e-6 / 4.20\%} & \textbf{5.41e-6 / 8.20\%} \\
\midrule
\multirow{4}{*}{\textit{Apple}} 
& Ground Truth & 4.16e-2  & 1.22e-5 & 1.22e-5 & 1.12e-5 \\
& Init(Intern-S1) & 1.30e-1 / 212.50\% & 6.50e-5 / 432.79\% & 6.50e-5 / 432.79\% & 6.50e-5 / 480.36\% \\
& Init(Gemini-3 Pro) & 7.00e-2 / 68.27\% & 4.33e-5 / 254.92\% & 4.33e-5 / 254.92\% & 3.66e-5 / 226.79\% \\
& Least-Square Baseline & 5.19e-2 / 24.76\% & 1.61e-2 / 31.97\% & 1.75e-2 / 43.44\% & 1.56e-2 / 39.29\%  \\
& Gradsim Baseline &4.39e-2 / 5.53\% & 1.59e-6 / 30.33\% & 1.38e-5 / 13.11\% & 1.35e-5 / 20.54\%   \\
& RigPI(Ours) & \textbf{4.23e-2 / 1.68\%} & \textbf{1.39e-5 / 13.93\%} & \textbf{1.30e-5 / 6.58\%} & \textbf{1.27e-5 / 12.29\%} \\
\midrule
\multirow{4}{*}{\textit{Drawer}} 
& Ground Truth & 1.11  & 1.25e-3 & 1.14e-2 & 1.02e-2 \\
& Init(Intern-S1) & 2.95 / 165.77\% & 3.56e-2 / 2748.00\% & 3.56e-2 / 2121.05\% & 6.02e-2 / 490.20\% \\
& Init(Gemini-3 Pro) & 1.60 / 44.14\% & 1.20e-2 / 860.00\% & 1.80e-2 / 57.89\% & 2.00e-2 / 96.08\% \\
& Least-Square Baseline &3.29 / 196.40\% & 1.07e-2 / 756.00\% & 2.94e-2 / 157.89\% & 2.88e-2 / 182.35\%  \\
& Gradsim Baseline &1.70 / 53.15\% & 1.77e-2 / 41.60\% & 1.82e-2 / 59.64\% & 1.61e-2 / 57.84\%  \\
& RigPI(Ours) & \textbf{1.19 / 7.21\%} & \textbf{1.39e-3 / 11.20\%} & \textbf{1.45e-2 / 27.19\%} & \textbf{1.22e-2 / 19.61\%} \\
\midrule
\multirow{4}{*}{\textit{Cabinet Door}} 
& Ground Truth & 8.72e-1  & 8.00e-3 & 8.00e-3 & 1.60e-2 \\
& Init(Intern-S1) & 9.80e-1 / 12.39\% & 1.00e-2 / 25.00\% & 1.31e-2 / 63.75\% & 2.30e-2 / 43.75\% \\
& Init(Gemini-3 Pro) & 7.70e-1 / 11.74\% & 7.00e-3 / 12.50\% & 7.00e-3 / 12.50\% & 1.40e-2 / 12.50\% \\
& Least-Square Baseline &1.49 / 70.87\% & 2.61e-2 / 226.25\% & 2.51e-2 / 213.75\% & 4.57e-2 / 185.63\%  \\
& Gradsim Baseline &1.13 / 29.59\% & 1.21e-2 / 51.25\% & 1.30e-2 / 62.50\% & 2.52e-2 / 57.50\%  \\
& RigPI(Ours) & \textbf{8.51e-1 / 2.41\%} & \textbf{7.67e-3 / 4.13\%} & \textbf{8.41e-3 / 5.13\%} & \textbf{1.86e-2 / 16.25\%} \\
\midrule
\multirow{4}{*}{\textit{Oven Door}} 
& Ground Truth & 6.42e-2  & 1.12e-4 & 3.62e-4 & 4.74e-4 \\
& Init(Intern-S1) & 1.25e-1 / 94.70\% & 3.50e-4 / 212.50\% & 4.50e-4 / 24.31\% & 1.03e-3 / 117.30\% \\
& Init(Gemini-3 Pro) & 8.00e-2 / 24.61\% & 1.40e-4 / 25.00\% & 5.00e-4 / 38.12\% & 5.20e-4 / 9.70\% \\
& Least-Square Baseline &8.22e-2 / 28.01\% & 2.64e-4 / 135.71\% & 3.15e-4 / 12.98\% & 5.63e-4 /18.78\%  \\
& Gradsim Baseline & 7.21e-1 / 12.31\% & 2.44e-4 / 117.86\% & 3.08e-4 / 14.92\% & 5.14e-4 / 8.44\% \\
& RigPI(Ours) & \textbf{6.67e-1 / 3.89\%} & \textbf{1.26e-4 / 12.50\%} & \textbf{3.28e-4 / 9.39\%} & \textbf{4.43e-4 / 6.54\%} \\
\bottomrule
\end{tabular}
\end{table*}

\vspace{-0.2cm}
\subsection{Evaluation Setup}
\label{subsec:task_setup}

\subsubsection{Accuracy and Stability of the Identification Process}  
Accuracy is measured by comparing estimated parameters with ground-truth (GT) values, while stability is quantified via the Coefficient of Variation (CoV) over five independent runs. High accuracy indicates close alignment with GT, and low CoV reflects robust and repeatable estimation. 

\subsubsection{Predictive Validity with Identified Parameters}
To evaluate the practical effectiveness of the identified parameters $\bm{\theta}^*$, we use them in simulation to predict object motion under novel, previously unseen applied wrenches. The predicted trajectories are then compared against real-world measurements. The trajectory reproduction error can assess how well the parameters generalize beyond the identification phase. 

\begin{table*}[t!]
\centering
\caption{Coefficient of Variation (\%) for identified parameters using different VLM priors. Lower values indicate higher stability across multiple runs. For the $\bm{com}$ vector, the CoV is averaged over its components, while for the Inertia tensor, the CoV is averaged over the principal diagonal elements ($I_{xx}, I_{yy}, I_{zz}$).}
\label{tab:identification_cov}
\begin{tabular}{@{}llcccc@{}}
\toprule
\textbf{Object} & \textbf{VLM Initializer} & \textbf{Mass CoV (\%)} & \textbf{$\bm{com}$ CoV (\%)} & \textbf{Inertia CoV (\%)} & \textbf{Friction CoV (\%)} \\
\midrule
\multirow{2}{*}{\textit{Cube}} & Intern-S1 &10.2 &25.7 &\textbf{18.7} &6.8 \\
 & Gemini-3 Pro &\textbf{7.4} &\textbf{17.2} &19.0 &\textbf{6.1} \\
\midrule
\multirow{2}{*}{\textit{Sphere}} & Intern-S1 &\textbf{8.4} &16.8 &17.5 &12.0 \\
 & Gemini-3 Pro &12.5 &\textbf{8.3} &\textbf{14.3} &\textbf{10.7} \\
\midrule
\multirow{2}{*}{\textit{Apple}} & Intern-S1 &9.1 &15.3 &14.2 &15.7 \\
 & Gemini-3 Pro &\textbf{7.3} &\textbf{11.0} &\textbf{11.4} &\textbf{7.3} \\
\midrule
\multirow{2}{*}{\textit{Drawer}} 
& Intern-S1 &9.4 &24.9 &\textbf{12.0} &14.2 \\
 & Gemini-3 Pro &\textbf{7.1} &\textbf{17.0} &15.2 &21.8 \\
\midrule
\multirow{2}{*}{\textit{Cabinet Door}} 
 & Intern-S1 &6.0 &\textbf{16.7} &22.4 &\textbf{20.7} \\
 & Gemini-3 Pro &\textbf{5.6} &19.9 &\textbf{19.2} &25.4 \\
 \midrule
\multirow{2}{*}{\textit{Oven Door}} 
 & Intern-S1 &8.1 &13.9 &\textbf{17.9} &23.7 \\
 & Gemini-3 Pro &\textbf{6.3} &\textbf{11.6} &19.2 &\textbf{15.7} \\
\bottomrule
\end{tabular}
\end{table*}

\section{RESULTS}
\label{sec:results}

\subsection{Accuracy and Stability of the Identification Process}

We evaluate RigPI's performance by analyzing the accuracy and stability of parameter identification.Table~\ref{tab:identification_accuracy} summarizes mass, center of mass ($\bm{com}$), principal moments of inertia ($I_{xx}, I_{yy}, I_{zz}$), and selected joint parameters for both unconstrained and multi-link rigid bodies. We compare GT values, initial guesses from two VLMs (Intern-S1 and Gemini-3 Pro), results from the traditional Least-Square baseline \cite{4048449} and the GradSim baseline \cite{gradsim}, and final parameters from RigPI. Our framework consistently outperforms the baseline, showing significant improvements in parameter accuracy. Figure~\ref{fig:RigPI_results} visualizes RigPI's output, displaying each object alongside predicted values. Some moments of inertia for multi-link rigid bodies are less accurate due to the limited range of joint motion; physically constrained movements make it difficult to excite all directions, challenging precise estimation.

For parameters that are difficult to measure precisely, such as friction coefficients and $\bm{com}$ locations, stability is further assessed via CoV across all objects (Table~\ref{tab:identification_cov}). Low CoV values indicate that RigPI achieves robust and reliable convergence.
\begin{table*}[t!] 
\centering 
\caption{Zero-shot estimation of physical parameters using Gemini 2.5 Pro and point cloud baseline. Each value shows the predicted value along (with relative error) with respect to the ground truth (GT). ‘Inertia’ refers to the average of the principal moments of inertia.} 
\label{tab:vlm_results_grid} 
\begin{minipage}[t]{0.2\linewidth}
\fbox{\begin{minipage}{\linewidth} 
\textbf{Prompt:} Please analyze the primary object in the provided image for its physical characteristics. First, extract its geometric properties: visualize its pose and infer its volume and internal structure (e.g., solid or hollow). Second, estimate its material properties, including the material density and the static coefficient of friction against the supporting surface. Finally, based strictly on these estimations, theoretically calculate the object's total mass, the location of its center of mass and the inertia tensor according to the center of mass. 
\end{minipage}} 
\end{minipage}%
\hfill 
\begin{minipage}[t]{0.75\linewidth} 
\begin{tabular}{@{}llcccc@{}}
\toprule 
\textbf{Object} & \textbf{Parameters} & \textbf{GT} & \textbf{VLM} & \textbf{Point Cloud} & \textbf{Rand.+Opt.}\\ 
\midrule 
\multirow{2}{*}{\textit{Cube}} 
& Mass (kg) & 2.00e-2 & \textbf{5.40e-2 / 170.0\%} & 6.25e-2 / 212.5\% & 6.15e-2 / 207.5\%\\ 
& Inertia & 5.3e-6 & 1.1e-5 / 107.5\% & 1.3e-5 / 145.3\% & \textbf{8.4e-6 / 58.5\%}\\ 
\midrule 
\multirow{2}{*}{\textit{Sphere}} 
& Mass & 2.00e-2 & 6.87e-2 / 243.5\% & 9.5e-2 / 375.0\% & \textbf{3.90e-2 / 95.0\%}\\ 
& Inertia & 5.0e-6 & \textbf{1.7e-5 / 240.0\%} & 5.9e-5 / 1080.0\% & 2.1e-5 / 320.0\%\\ 
\midrule 
\multirow{2}{*}{\textit{Apple}} 
& Mass & 4.16e-2 & 6.87e-2 / 65.1\% & 1.09e-1 / 162.0\% & \textbf{5.27e-2 / 26.7\%}\\ 
& Inertia & 1.2e-5 & 4.1e-5 / 241.7\% & 5.7e-5 / 375.0\% & \textbf{3.2e-5 / 166.7\%}\\ 
\midrule 
\multirow{2}{*}{\textit{Wooden Drawer}} 
& Mass & 1.11 & \textbf{1.60 / 44.1\%} & 1.88 / 69.4\% & 1.81 / 63.1\%\\ 
& Inertia & 7.6e-3 & 1.7e-2 / 123.7\% & 1.7e-2 / 123.7\% & \textbf{1.6e-2 / 110.5\%}\\ 
\midrule 
\multirow{2}{*}{\textit{Cabinet Door}} 
& Mass & 8.72e-1 & \textbf{7.70e-1 / 11.7\%} & 1.67 / 91.3\% & 9.77e-2 / 12.0\%\\ 
& Inertia & 1.1e-2 & \textbf{9.3e-3 / 15.5\%} & 4.5e-2 / 309.1\% & 3.2e-2 / 190.91\%\\ 
\midrule 
\multirow{2}{*}{\textit{Oven Door}} 
& Mass & 6.42e-2 & \textbf{8.00e-2 / 24.6\%} & 1.29 / 100.1\% & 8.95e-2 / 39.4\%\\ 
& Inertia & 3.2e-4 & 3.9e-4 / 21.9\% & 5.7e-3 / 78.1\% & \textbf{3.7e-4 / 15.6\%}\\ 
\bottomrule 
\end{tabular} 
\end{minipage} 
\end{table*}

\subsection{Predictive Validity with Identified Parameters}
\label{subsec:trajectory_reproduction_and_validation}

To demonstrate the impact of identified parameters $\bm{\theta}^*$, we simulate each object’s response to novel robot-applied forces and torques. Predicted trajectories are compared with real robot measurements to evaluate trajectory reproduction.

\begin{figure}[t!]
    \centering
    \includegraphics[width=0.9\columnwidth]{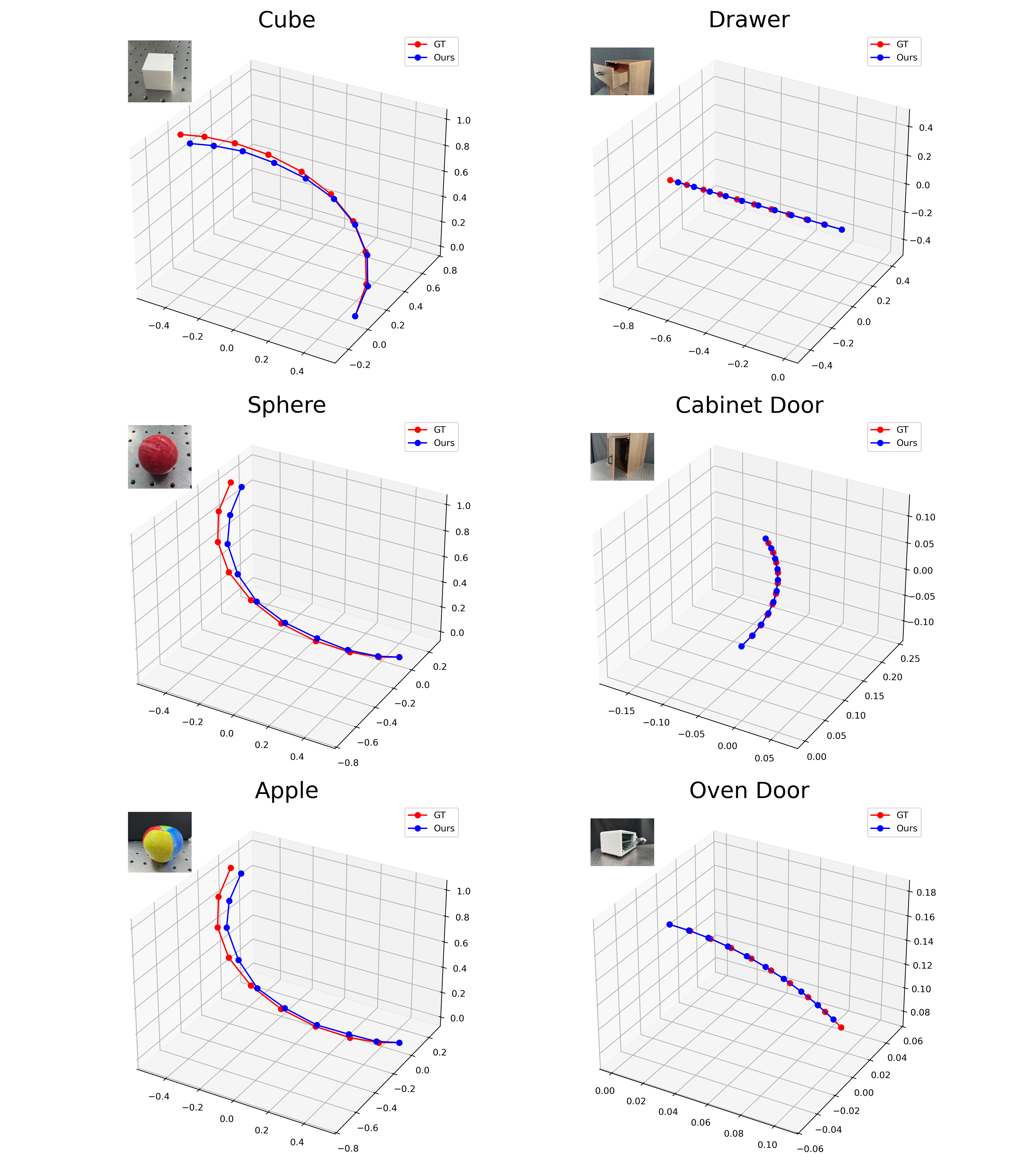}
    \caption{
    Qualitative results of trajectory reproduction by predictive validity with identified parameters. 
    Simulated trajectories (blue) closely follow the ground-truth (red). 
    }
    \label{fig:trajectory_reproduction}
    \vspace{-0.5cm}
\end{figure}

We quantitatively assess the accuracy of the identified parameters by computing the loss on new trajectories using equation \eqref{equa:loss function}. As Table~\ref{tab:trajectory_error} shows, low loss values across all objects indicate accurate prediction of responses under novel dynamics. Values are averaged over five trajectories. Representative predicted trajectories in Fig.~\ref{fig:trajectory_reproduction} illustrate how the identified parameters capture each object’s dynamics.




\begin{table}[h!]
\centering
\caption{Quantitative evaluation of trajectory reproduction by predictive validity with identified parameters.}
\label{tab:trajectory_error}
\begin{tabular}{@{}lccc@{}}
\toprule
\textbf{Object} & \textbf{Reproduction Loss ($\mathcal{L}_{\text{pose}}$)}   & \textbf{Endpt Dist.(m)} \\
\midrule
\textit{Cube} & \textit{9.98e-2} & \textit{8.25e-2}\\
\textit{Sphere} & \textit{2.93e-1} & \textit{4.53e-2}\\
\textit{Apple} & \textit{1.30e-1} & \textit{5.46e-2}\\
\textit{Drawer} & \textit{4.41e-1} & \textit{3.79e-2}\\
\textit{Cabinet Door} & \textit{3.94e-1} & \textit{5.04e-2}\\
\textit{Oven Door} & \textit{4.38e-1} & \textit{4.21e-2}\\
\bottomrule
\end{tabular}
\vspace{-0.3cm}
\end{table}

\subsection{Ablation Study}
\label{subsec:ablation}
\subsubsection{Effectiveness of VLM Priors}

RigPI relies on VLMs to provide reasonable initial guesses and a valid search space. We evaluate zero-shot VLM estimation on diverse objects using a single RGB-D image per object, where the VLM predicts geometric and material properties, which are used to derive mass and inertia, and compared with ground-truth.

We compare it with other initialization methods: randomized initialization (``Rand.''), geometry-aware initialization (``Point Cloud''), and the ground truth.

For randomized initialization, we assign mass, center of mass, and inertia values by sampling uniformly within a given range. For instance, mass is sampled between 0 and the maximum payload, the center of mass is sampled within the workspace bounds (e.g., the tabletop), and the inertia tensor is computed based on these sampled values. For geometry-aware initialization, we reconstruct a point cloud from the RGB image and approximate it by a convex hull. Assuming uniform density, mass and inertia are computed analytically. While this geometry-based method can be effective for near-ideal shapes, its accuracy degrades for irregular or concave objects due to reconstruction errors.

As shown in Table~\ref{tab:vlm_results_grid}, all the results are initialized by different methods and optimized later. The VLM produces estimates much closer to ground truth, yielding more reliable initial guesses $\bm{\theta}^{(0)}$. 
As for the randomized initialization, ``Rand. + Opt.'' shows competitive results with ``VLM''. However, randomized initialization can easily lead to rapidly increasing and unstable loss, whereas VLM guidance maintains low and consistent loss. The seemingly good results of ``Rand. + Opt.'' are primarily due to convergence to physically invalid local minima, where numerical values can become extremely large and optimization unreliable. Unlike other initialization methods, which consistently produce reasonable outcomes, random initialization exhibits high variability across trials: some runs converge well, while others diverge dramatically to extreme and physically incorrect values. Such cases are excluded from the results reported in the table.

\subsubsection{Effect of Dynamic Learning Rate $\alpha$}
\begin{figure}[t]
    \centering
    \includegraphics[width=0.4\textwidth]{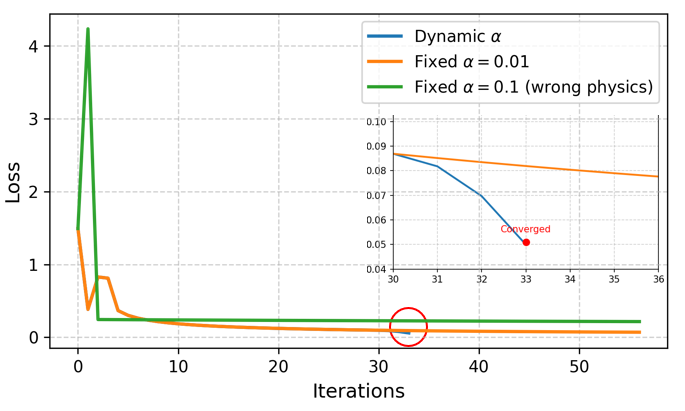} 
    \caption{Illustrative example of convergence curves for different learning rate strategies. 
    The small fixed $\alpha$ results in slow convergence, whereas the large fixed $\alpha$ may lead 
    to physically invalid solutions. The dynamic $\alpha$ strategy adaptively balances the update magnitude, 
    ensuring fast and stable convergence.}
    \label{fig:alpha_curve}
    \vspace{-0.5cm}
\end{figure}

Different schemes for setting $\alpha$ could also influence the optimization. 
We compare three different $\alpha$ settings: 0.01, 0.1, and a dynamic setting.
A properly chosen $\alpha$ prevents updates from being too small, causing slow convergence, or too large, risking physically invalid parameters. Figure~\ref{fig:alpha_curve} illustrates loss curves under different $\alpha$ settings, confirming that dynamic $\alpha$ primarily regulates step size for safe and efficient optimization rather than tuning final accuracy.

\section{CONCLUSION AND FUTURE WORKS}

In this work, we present RigPI, a robust solution to the ill-posed problem of dynamic parameter identification. Our approach anchors differentiable physics simulation with vision-guided semantic priors, transforming the task into a tractable optimization problem. Experiments show that RigPI yields accurate and stable estimates of inertial and frictional properties for diverse unconstrained and multi-link rigid objects. Integrating prior knowledge with physical reasoning offers a promising path toward physically-aware robotic systems capable of complex manipulation. Future work will focus on autonomous exploration for parameter discovery and integrating these models into planning and control frameworks to further enhance robotic capabilities.



\addtolength{\textheight}{-12cm}   









\bibliographystyle{IEEEtran}
\bibliography{bibtex}

@article{hu2019difftaichi,
  title={DiffTaichi: Differentiable Programming for Physical Simulation},
  author={Hu, Yuanming and Anderson, Luke and Li, Tzu-Mao and Sun, Qi and Carr, Nathan and Ragan-Kelley, Jonathan and Durand, Fr{\'e}do},
  journal={International Conference on Learning Representations (ICLR)},
  year={2020}
}

@article{howelllecleach2022,
	title={Dojo: A Differentiable Physics Engine for Robotics},
	author={Howell, Taylor and Le Cleac'h, Simon and Bruedigam, Jan and Kolter, Zico and Schwager, Mac and Manchester, Zachary},
	journal={arXiv preprint arXiv:2203.00806},
	url={https://arxiv.org/abs/2203.00806},
	year={2022}
}

@article{lidec2021,
  author={Le Lidec, Quentin and Kalevatykh, Igor and Laptev, Ivan and Schmid, Cordelia and Carpentier, Justin},
  journal={IEEE Robotics and Automation Letters}, 
  title={Differentiable Simulation for Physical System Identification}, 
  year={2021},
  volume={6},
  number={2},
  pages={3413-3420},
  doi={10.1109/LRA.2021.3062323}}

@misc{gradsim,
      title={gradSim: Differentiable simulation for system identification and visuomotor control}, 
      author={Krishna Murthy Jatavallabhula and Miles Macklin and Florian Golemo and Vikram Voleti and Linda Petrini and Martin Weiss and Breandan Considine and Jerome Parent-Levesque and Kevin Xie and Kenny Erleben and Liam Paull and Florian Shkurti and Derek Nowrouzezahrai and Sanja Fidler},
      year={2021},
      eprint={2104.02646},
      archivePrefix={arXiv},
      primaryClass={cs.CV},
      url={https://arxiv.org/abs/2104.02646}, 
}

@misc{bai2025interns1scientificmultimodalfoundation,
  title={Intern-S1: A Scientific Multimodal Foundation Model},
  author={Lei Bai, Zhongrui Cai, Yuhang Cao {\em et al.}},
  year={2025},
  eprint={2508.15763},
  archivePrefix={arXiv},
  primaryClass={cs.LG},
  url={https://arxiv.org/abs/2508.15763},
}

@misc{comanici2025gemini25pushingfrontier,
  title={Gemini 2.5: Pushing the Frontier with Advanced Reasoning, Multimodality, Long Context, and Next Generation Agentic Capabilities}, 
  author={Gheorghe Comanici, Eric Bieber, Mike Schaekermann {\em et al.}},
  year={2025},
  eprint={2507.06261},
  archivePrefix={arXiv},
  primaryClass={cs.CL},
  url={https://arxiv.org/abs/2507.06261}, 
}

@misc{Toussaint2018DifferentiablePA,
  title={Differentiable Physics and Stable Modes for Tool-Use and Manipulation Planning},
  author={Marc Toussaint and Kelsey R. Allen and Kevin A. Smith and Joshua B. Tenenbaum},
  booktitle={Robotics: Science and Systems},
  year={2018},
  url={https://api.semanticscholar.org/CorpusID:46980516}
}

@article{le_cleach_differentiable_2023,
  title = {Differentiable physics simulation of dynamics-augmented neural objects},
  volume = {8},
  number = {5},
  journal = {IEEE Robotics and Automation Letters},
  author = {Le Cleac'h, Simon and Yu, Hong-Xing and Guo, Michelle and Howell, Taylor and Gao, Ruohan and Wu, Jiajun and Manchester, Zachary and Schwager, Mac},
  year = {2023},
  note = {Publisher: IEEE},
  pages = {2780--2787}
}

@misc{degrave2018differentiablephysicsenginedeep,
      title={A Differentiable Physics Engine for Deep Learning in Robotics}, 
      author={Jonas Degrave and Michiel Hermans and Joni Dambre and Francis wyffels},
      year={2018},
      eprint={1611.01652},
      archivePrefix={arXiv},
      primaryClass={cs.NE},
      url={https://arxiv.org/abs/1611.01652}, 
}

@misc{chen2025learningobjectpropertiesusing,
      title={Learning Object Properties Using Robot Proprioception via Differentiable Robot-Object Interaction}, 
      author={Peter Yichen Chen and Chao Liu and Pingchuan Ma and John Eastman and Daniela Rus and Dylan Randle and Yuri Ivanov and Wojciech Matusik},
      year={2025},
      eprint={2410.03920},
      archivePrefix={arXiv},
      primaryClass={cs.RO},
      url={https://arxiv.org/abs/2410.03920}, 
}

@misc{sanchezgonzalez2018graphnetworkslearnablephysics,
      title={Graph networks as learnable physics engines for inference and control}, 
      author={Alvaro Sanchez-Gonzalez and Nicolas Heess and Jost Tobias Springenberg and Josh Merel and Martin Riedmiller and Raia Hadsell and Peter Battaglia},
      year={2018},
      eprint={1806.01242},
      archivePrefix={arXiv},
      primaryClass={cs.LG},
      url={https://arxiv.org/abs/1806.01242}, 
}

@misc{lutter2019deeplagrangiannetworksusing,
      title={Deep Lagrangian Networks: Using Physics as Model Prior for Deep Learning}, 
      author={Michael Lutter and Christian Ritter and Jan Peters},
      year={2019},
      eprint={1907.04490},
      archivePrefix={arXiv},
      primaryClass={cs.LG},
      url={https://arxiv.org/abs/1907.04490}, 
}

@inproceedings{wu2015nips,
 author = {Wu, Jiajun and Yildirim, Ilker and Lim, Joseph J and Freeman, Bill and Tenenbaum, Josh},
 booktitle = {Advances in Neural Information Processing Systems},
 editor = {C. Cortes and N. Lawrence and D. Lee and M. Sugiyama and R. Garnett},
 pages = {},
 publisher = {Curran Associates, Inc.},
 title = {Galileo: Perceiving Physical Object Properties by Integrating a Physics Engine with Deep Learning},
 url = {https://proceedings.neurips.cc/paper_files/paper/2015/file/d09bf41544a3365a46c9077ebb5e35c3-Paper.pdf},
 volume = {28},
 year = {2015}
}

@InProceedings{pmlr-v78-standley17a,
  title = 	 {image2mass: Estimating the Mass of an Object from Its Image},
  author = 	 {Standley, Trevor and Sener, Ozan and Chen, Dawn and Savarese, Silvio},
  booktitle = 	 {Proceedings of the 1st Annual Conference on Robot Learning},
  pages = 	 {324--333},
  year = 	 {2017},
  editor = 	 {Levine, Sergey and Vanhoucke, Vincent and Goldberg, Ken},
  volume = 	 {78},
  series = 	 {Proceedings of Machine Learning Research},
  month = 	 {13--15 Nov},
  publisher =    {PMLR},
  url = 	 {https://proceedings.mlr.press/v78/standley17a.html}
}

@article{chow2025physbench,
  title={PhysBench: Benchmarking and Enhancing Vision-Language Models for Physical World Understanding},
  author={Chow, Wei and Mao, Jiageng and Li, Boyi and Seita, Daniel and Guizilini, Vitor and Wang, Yue},
  journal={arXiv preprint arXiv:2501.16411},
  year={2025}
}

@misc{brohan2023rt2visionlanguageactionmodelstransfer,
      title={RT-2: Vision-Language-Action Models Transfer Web Knowledge to Robotic Control}, 
      author={Anthony Brohan, Noah Brown, Justice Carbajal {\em et al.}},
      year={2023},
      eprint={2307.15818},
      archivePrefix={arXiv},
      primaryClass={cs.RO},
      url={https://arxiv.org/abs/2307.15818}, 
}

@misc{lai2024visionlanguagemodelbasedphysicalreasoning,
      title={Vision-Language Model-based Physical Reasoning for Robot Liquid Perception}, 
      author={Wenqiang Lai and Yuan Gao and Tin Lun Lam},
      year={2024},
      eprint={2404.06904},
      archivePrefix={arXiv},
      primaryClass={cs.RO},
      url={https://arxiv.org/abs/2404.06904}, 
}

@misc{zhou2025physvlmenablingvisuallanguage,
      title={PhysVLM: Enabling Visual Language Models to Understand Robotic Physical Reachability}, 
      author={Weijie Zhou and Manli Tao and Chaoyang Zhao and Haiyun Guo and Honghui Dong and Ming Tang and Jinqiao Wang},
      year={2025},
      eprint={2503.08481},
      archivePrefix={arXiv},
      primaryClass={cs.RO},
      url={https://arxiv.org/abs/2503.08481}, 
}

@misc{elnoor2024robotnavigationusingphysically,
      title={Robot Navigation Using Physically Grounded Vision-Language Models in Outdoor Environments}, 
      author={Mohamed Elnoor and Kasun Weerakoon and Gershom Seneviratne and Ruiqi Xian and Tianrui Guan and Mohamed Khalid M Jaffar and Vignesh Rajagopal and Dinesh Manocha},
      year={2024},
      eprint={2409.20445},
      archivePrefix={arXiv},
      primaryClass={cs.RO},
      url={https://arxiv.org/abs/2409.20445}, 
}

@INPROCEEDINGS{4048449,
  author={Atkeson, Chirstopher G. and An, Chae H. and Hollerbach, John M.},
  booktitle={1985 24th IEEE Conference on Decision and Control}, 
  title={Rigid body load identification for manipulators}, 
  year={1985},
  volume={},
  number={},
  pages={996-1002},
  doi={10.1109/CDC.1985.268649}}

@INPROCEEDINGS{4650672,
  author={Kubus, Daniel and Kroger, Torsten and Wahl, Friedrich M.},
  booktitle={2008 IEEE/RSJ International Conference on Intelligent Robots and Systems}, 
  title={On-line estimation of inertial parameters using a recursive total least-squares approach}, 
  year={2008},
  volume={},
  number={},
  pages={3845-3852},
  doi={10.1109/IROS.2008.4650672}}

@article{Xu_Fan_Fang_Zhu_Zhao_2022, 
title={An accurate identification method based on double weighting for inertial parameters of robot payloads}, volume={40}, DOI={10.1017/S0263574722000960}, number={12}, journal={Robotica}, author={Xu, Tian and Fan, Jizhuang and Fang, Qianqian and Zhu, Yanhe and Zhao, Jie}, year={2022}, pages={4358–4374}}

@software{newton,
  title = {{Newton}: {GPU}-accelerated physics simulation for robotics, and simulation research.},
  author = {{Newton Contributors}},
  year = {2025},
  url = {https://github.com/newton-physics/newton},
  organization = {{Newton a Series of LF Projects, LLC}},
  license = {Apache-2.0}
}

@inproceedings{diffstir,
  title = {Differentiable Fluid Physics Parameter Identification By Stirring and For Stirring},
  author = {Xu*, Wenqiang and Zheng*, Dongzhe and Li, Yutong and Ren, Jieji and Lu, Cewu},
  booktitle = {{IEEE/RSJ} International Conference on Intelligent Robots and Systems},
  pages = {},
  year = {2024},
  organization = {IEEE},
}

@article{diffcp,
  title = {Differentiable Cloth Parameter Identification and State Estimation in Manipulation},
  author = {Zheng*, Dongzhe and Yao*, Siqiong and Xu, Wenqiang and Lu, Cewu},
  journal = {IEEE Robotics and Automation Letters},
  volume = {},
  number = {},
  pages = {},
  year = {2024},
  publisher = {IEEE},
}

\end{document}